\documentclass[final]{cvpr}

\usepackage{times}
\usepackage{epsfig}
\usepackage{graphicx}
\usepackage{amsmath}
\usepackage{amssymb}
\usepackage{graphicx}
\usepackage{booktabs}

\usepackage{listings}
\usepackage{xcolor}

\definecolor{codegreen}{rgb}{0,0.6,0}
\definecolor{codegray}{rgb}{0.5,0.5,0.5}
\definecolor{codepurple}{rgb}{0.58,0,0.82}
\definecolor{backcolour}{rgb}{0.95,0.95,0.95}

\lstdefinestyle{mystyle}{
  backgroundcolor=\color{backcolour},   commentstyle=\color{codegreen},
  keywordstyle=\color{magenta},
  numberstyle=\tiny\color{codegray},
  stringstyle=\color{codepurple},
  basicstyle=\ttfamily\small,
  breakatwhitespace=false,         
  breaklines=true,                 
  captionpos=b,                    
  keepspaces=true,                 
  numbers=left,                    
  numbersep=5pt,                  
  showspaces=false,                
  showstringspaces=false,
  showtabs=false,                  
  tabsize=2
}
\usepackage{soul}
\usepackage{color}

\usepackage[pagebackref=true,breaklinks=true,colorlinks,bookmarks=false]{hyperref}

\def\cvprPaperID{4229} 

\def\confYear{CVPR 2021}

\title{Weakly-supervised Instance Segmentation via Class-agnostic Learning with Salient Images}

\begin{document}

\author{Xinggang Wang$^{\dag}$ \ \  Jiapei Feng$^{\dag}$ \ \  Bin Hu$^{\dag}$ \ \   Qi Ding$^{\ddag}$ \ \ Longjin Ran$^{\ddag}$ \ \ Xiaoxin Chen$^{\ddag}$ \ \  Wenyu Liu$^{\dag\star}$  \\
        $^{\dag}$Huazhong University of Science and Technology \ \   $^{\ddag}$VIVO Inc. \\    
        {\tt\small \{xgwang,fjp,binhu\_19,liuwy\}@hust.edu.cn \{qi.ding,lj.ran,xiaoxin.chen\}@vivo.com}
}

\maketitle

\begin{abstract}
    Humans have a strong class-agnostic object segmentation ability and can outline boundaries of unknown objects precisely, which motivates us to propose a box-supervised class-agnostic object segmentation (BoxCaseg) based solution for weakly-supervised instance segmentation. The BoxCaseg model is jointly trained using box-supervised images and salient images in a multi-task learning manner. The fine-annotated salient images provide class-agnostic and precise object localization guidance for box-supervised images. The object masks predicted by a pretrained BoxCaseg model are refined via a novel merged and dropped strategy as proxy ground truth to train a Mask R-CNN for weakly-supervised instance segmentation. Only using $7991$ salient images, the weakly-supervised Mask R-CNN is on par with fully-supervised Mask R-CNN on PASCAL VOC and significantly outperforms previous state-of-the-art box-supervised instance segmentation methods on COCO. The source code, pretrained models and datasets are available at \url{https://github.com/hustvl/BoxCaseg}.
\end{abstract}

\let\thefootnote\relax\footnotetext{$^\star$Wenyu Liu is the corresponding author.}

\vspace{-3mm}
\section{Introduction}
Weakly-supervised instance segmentation (WSIS) is important in computer vision for at least two reasons: First, it reduces human annotation efforts so that it is useful in building label-efficient visual recognition systems. Second, humans can learn to segment objects without perfect/exact labels, and building a visual learning system similar to human perception is a long-term goal of computer vision; thus the research weakly-supervised instance segmentation is a way to approach human perception. In this paper, we investigate the problem of weakly-supervised instance segmentation using bounding box annotations rather than the precise mask annotations. 

\begin{figure}[thp]
\centering
\includegraphics[width=1\linewidth]{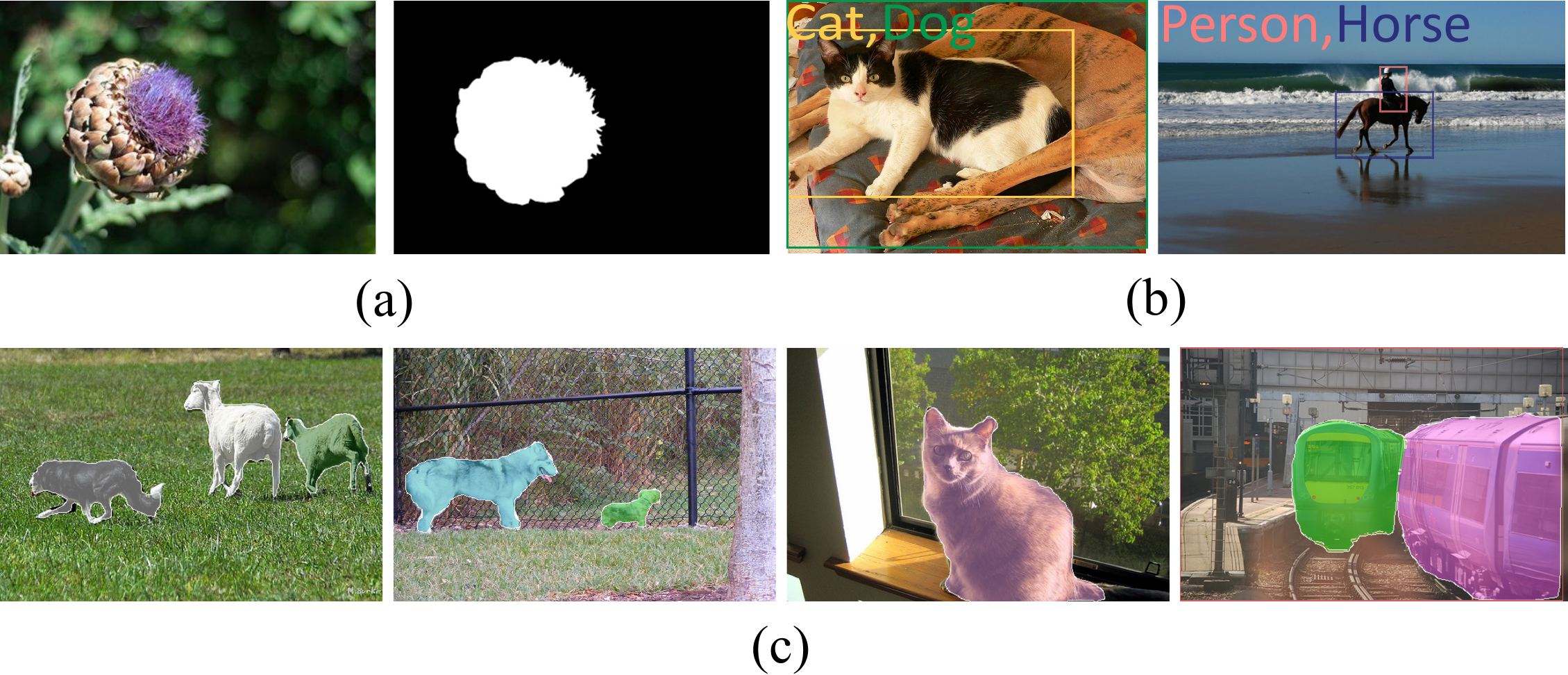}
\caption{The top row shows a salient training image (a) and two box-supervised training images (b). Our model is jointly trained using both salient images and box-supervised images and obtains high quality instance segmentation results (c).}
\label{fig:first}
\end{figure}

As reported in \cite{bearman2016s}, labeling a mask takes about $79$ seconds while labeling a bounding box takes only about $10$ seconds on average. Thus, if a WSIS method can obtain the segmentation precision on par with the state-of-the-art fully supervised instance segmentation methods (\eg, Mask R-CNN \cite{he2017mask}), it is extremely useful and has great potential to be applied in real world applications. To solve this problem, previous methods have applied unsupervised image segmentation methods based on hand-crafted features \cite{rother2004grabcut,pont2016multiscale}, pseudo labeling \cite{dai2015boxsup,song2019box,arun2020weakly}, multi-instance learning \cite{hsu2019weakly} and attention networks \cite{kulharia12356box2seg} in various WSIS frameworks.

\begin{figure*}[htp]
\centering
\includegraphics[width=1\textwidth]{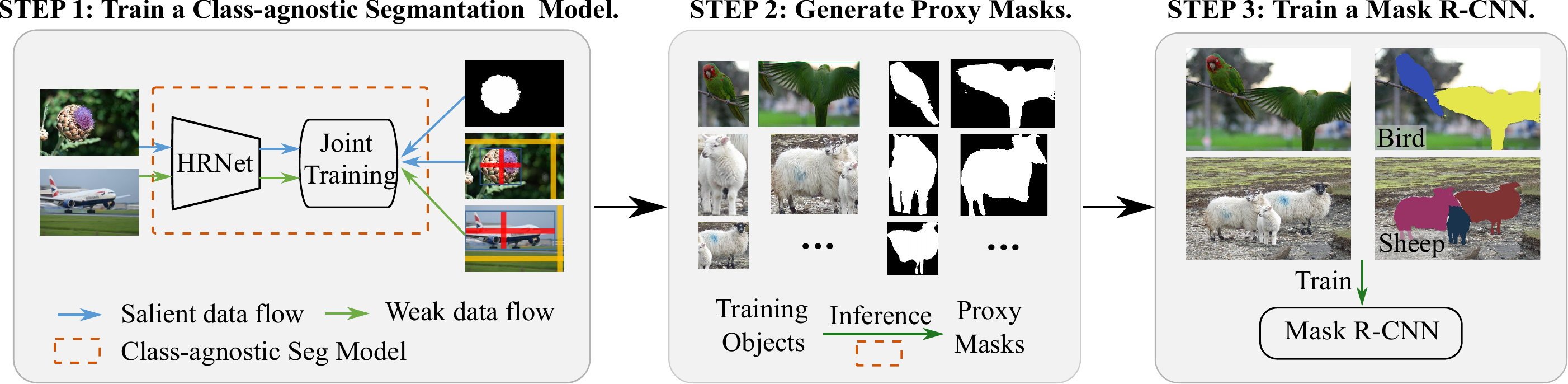}
\caption{The pipeline of our approach. Our training process is divided into three steps. First, we obtain a class-agnostic segmentation model (BoxCaseg) through the joint training of box-supervised dataset and salient object segmentation dataset (Sec.~\ref{Sec:Joint_train}). Then, we use the class-agnostic segmentation model to produce high quality predictions for training instances. Those segmentation predictions are merged to generate proxy instance masks for training images. Finally, we re-train a Mask R-CNN with proxy masks (Sec.~\ref{sec:retraining}).}
\label{fig:framework}
\end{figure*}

We think the key problem in WSIS is box-supervised class-agnostic object segmentation (BoxCaseg), \ie, given the object bounding box, we need to infer pixel-level object mask. Humans have strong ability on the task of BoxCaseg and can perceive object boundaries even the objects are from unknown semantic categories. To mimic this class-agnostic human learning process, we propose a joint training scheme, in which a salient object segmentation (also known as salient object detection) dataset that contains class-agnostic pixel-level object location information is applied to simulate humans' historical memory. The salient images are trained with box-annotated images (as shown in Fig.~\ref{fig:first}) in a single deep multi-task network and are mixed in the same mini-batches. The tasks include pixel-level labeling salient images and multi-instance classification \cite{hsu2019weakly} of both box-supervised images and salient images. During training, salient images can provide salient knowledge to guide the segmentation of box-supervised images. Besides, inspired by Mask$^{\star}$ R-CNN \cite{hu2018learning}, weight transfer is performed from the multi-instance learning task to the pixel-labeling task to enhance the joint learning.

The learned BoxCaseg model is applied to generate object masks for box-supervised training images. The masks are refined by our proposed merging and dropping strategies for training a Mask R-CNN model for WSIS. Experiments are carried out on the PASCAL VOC and COCO datasets. At last, our main contributions are highlighted as follows. 
\vspace{-2mm}
\begin{enumerate}
\item We propose a novel weakly-supervised instance segmentation method based on box-supervised class-agnostic object segmentation, in which class-agnostic precise salient object localization information is utilized as auxiliary memory to promote weakly-supervised instance segmentation during training.
\vspace{-2mm}
\item We propose new mask merging and dropping strategies to obtain high-quality proxy masks for pseudo-training Mask R-CNN.
\vspace{-2mm}
\item The class-agnostic segmentation model has great generalization ability. Only using $7991$ salient images which are disjointed with PASCAL and COCO, for the first time, a box-supervised instance segmentation obtains similar performance with its fully-supervised counterpart. On COCO, our result is significantly better than previous state-of-the-art WSIS methods.
\end{enumerate}

\vspace{-3mm}
\section{Related Work}
\vspace{-1mm}

Instance segmentation acquires fine-grained but expensive pixel-level annotations while many weakly supervised methods can use only coarser annotations, such as image tags \cite{ahn2019weakly,cholakkal2019object,ge2019label,laradji2019masks,zhou2018weakly,zhu2019learning}, points \cite{laradji2019instance}, scribble\cite{lin2016scribblesup,tang2018normalized}, and bounding boxes \cite{arun2020weakly,hsu2019weakly,khoreva2017simple, tian2020boxinst, bbam}.

Our setup considers using object bounding box as the weak annotation. To obtain the segmentation from the bounding box, there is a kind of methods \cite{dai2015boxsup,papandreou2015weakly,khoreva2017simple} utilizing traditional image segmentation algorithms to segment objects, such as CRF \cite{krahenbuhl2011efficient}, GrabCut \cite{rother2004grabcut} and MCG \cite{pont2016multiscale}. However, these methods rely on low-level image features and cannot obtain stable WSIS performance. Dai \etal \cite{dai2015boxsup} first proposes to combine deep networks with hand-crafted object proposals for WSIS via iterative training and pseudo labeling. Then methods such as \cite{song2019box,kulharia12356box2seg} focus on the refinement of pseudo mask with the mean filling rates. Arun \etal. \cite{arun2020weakly} propose a conditional network and a prediction network which supervise each other. Unlike the above methods, BBTP \cite{hsu2019weakly} designs a multi-instance learning (MIL) formulation to train a weakly-supervised segmentation model. On the basis of BBTP, BoxInst \cite{tian2020boxinst} employs the projection loss to replace the MIL loss and improves the pairwise loss with the prior derived from proximal pixels’ colors. Our weak segmentation head is also based on BBTP and does not rely on any traditional algorithms.

Our method is also based on the development of salient object detection (SOD, as known as salient object segmentation) \cite{jiang2013salient,yang2013saliency,li2013saliency,borji2015salient,li2016deep,hou2017deeply,wang2017learning}, which aims at finding visually attractive objects in an image and segmenting them as a binary mask. It has achieved great success based on deep learning. Recent deep SOD methods are beneficial from prediction erasing via multi-level feature fusion \cite{zhang2017amulet}, reverse attention \cite{chen2018reverse}, global and local context integration \cite{Liu2018PiCANet}, cascaded partial decoder \cite{wu2019cascaded}, edge-guided network \cite{zhao2019egnet}, pooling-based global guidance \cite{liu2019simple} and nested U-Net (\ie, U$^2$Net) \cite{qin2020u2}.

\begin{figure}[ht]
\centering
\includegraphics[width=1\linewidth]{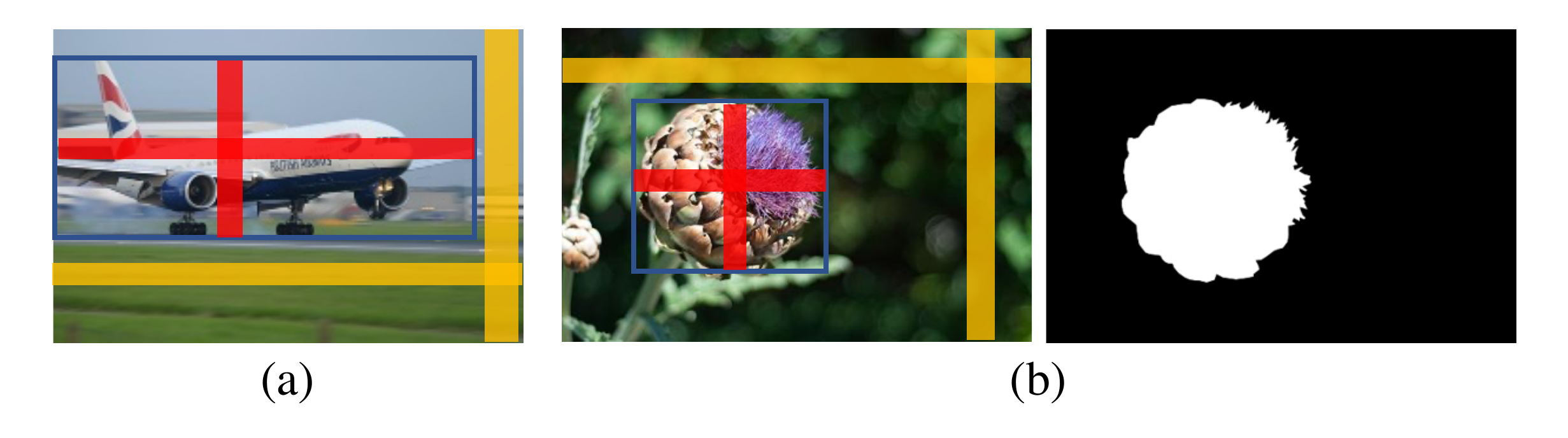}
\caption{Two examples from box-supervised dataset (a) and salient object segmentation dataset (b). According to ground truth box (blue boxes in images), we can split each horizontal and vertical crossing lines of images into positive bags and negative bags. Red lines and yellow lines represent positive bags and negative bags respectively.}
\label{fig:bags}
\end{figure}

Salient object segmentation information had been applied as a cue for weakly-supervised instance/semantic segmentation, \eg, \cite{huang2018weakly, fanECCV18GP, zeng2019joint, zhou2020learning}, in which \cite{zeng2019joint,zhou2020learning} are the most related works to our method. In \cite{zeng2019joint} salient images are jointly trained with images with category labels for class-specific semantic segmentation and saliency maps are obtain via aggregating the class-specific prediction results. Conceptually different from \cite{zeng2019joint}, our method is for class-agnostic object segmentation and does not depend on the category intersection between salient images and box-supervised images. ShapeProp \cite{zhou2020learning} performs pixel-wise MIL and message passing within bounding boxes to obtain class-agnostic saliency information as prior for semi-supervised instance segmentation. In contrast, our motivation is to simulate human’s strong class-agnostic object segmentation ability and the domain gap between salient images and weakly-supervised images is much larger than that in ShapeProp. Besides, there are weakly-supervised learning methods for salient object segmentation, \eg, \cite{li2018weakly,zeng2019multi,zhang2020weakly}. In our work, the salient object information is utilized as an auxiliary memory for weakly-supervised segmenting generic images.


At last, class-agnostic object segmentation is becoming a hot research topic. Pinheiro \etal \cite{pinheiro2015learning} first studies the problem of class-agnostic object candidate segmentation. Zhang \etal \cite{zhang2019canet} perform class-agnostic segmentation via dense comparison for few-shot object segmentation. Sharma \etal \cite{sharma2020class} propose a novel class-agnostic segmentation loss function for better salient object detection and segmentation. Our work shows that class-agnostic learning is extremely helpful to WSIS.


\vspace{-1mm}
\section{Method}
\vspace{-1mm}



To describe the proposed method in a self-contained manner, we first introduce the MIL-based WSIS method \cite{hsu2019weakly}, then formulate the proposed joint training scheme for BoxCaseg, and finally describe how to train a Proxy Mask R-CNN. The overall pipeline of our method is illustrated in Fig.~\ref{fig:framework}.

\subsection{Weakly-supervised Instance Segmentation via Multi-instance Learning}
\label{sec:bbtp}

In order to approach the instance segmentation task with bounding box annotations, Hsu \etal \cite{hsu2019weakly} propose a MIL-based method that utilizes the bounding box tightness prior (BBTP). The method replaces the original segmentation loss with a novel MIL loss in Mask R-CNN \cite{he2017mask}. 
As shown in Fig.~\ref{fig:bags}, the entire region of the object is in the bounding box and each row or each column in the bounding box must contain at least one pixel belonging to the object. BBTP regards these rows and columns as positive bags. Other rows (or columns) outside the bounding box are treated as negative bags. A bag contains multiple pixels/features on the row (or column). As a result, either given the bounding box from box-supervised images or the box generated by salient object mask, we can split the image into positive bags and negative bags. 

In Mask R-CNN, it is worth to note that there are multiple region proposals corresponding to one object. Thus the positive and negative bags generation process will be performed for each region proposal separately. Once the positive and negative bags are generated, its classification probabilities are generated by max-pooling pixel classification probabilities. The bag classification loss can be back-propagated to optimize pixel classifier to achieve pixel segmentation.



\begin{figure*}[ht]
\centering
\includegraphics[width=0.9\linewidth]{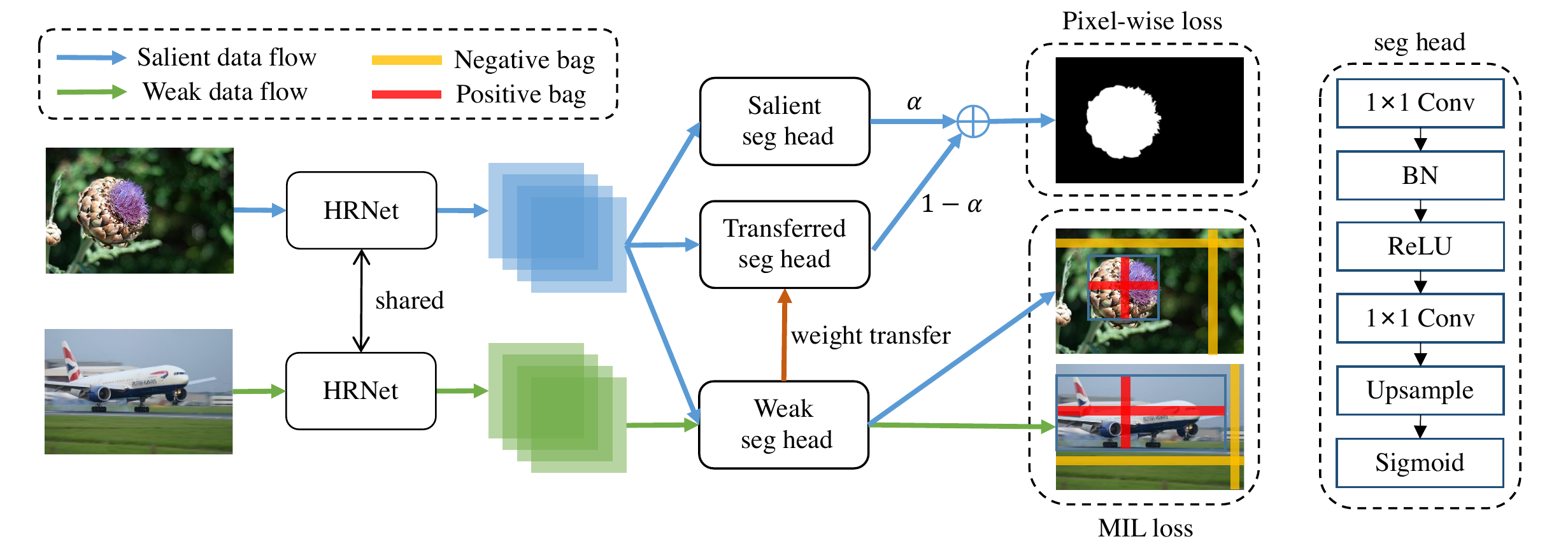}
\caption{The joint training scheme for box-supervised class-agnostic object segmentation. A salient image and a box-supervised image within a mini-batch are fed into a backbone network (\ie, HRNet \cite{wang2020deep}) to extract convolutional feature maps. The weak convolutional feature map is then fed into the weak segmentation head trained with the MIL loss described in Eq.~\eqref{eq:mil}. The salient convolutional feature map is simultaneously fed into the weak segmentation head, the transferred segmentation head, and the salient segmentation head. The last two heads are fused and supervised via a pixel-labeling loss (as described in Eq.~\eqref{eq:pix}).}
\label{fig:caseg}
\end{figure*}

\subsection{Joint Training with Salient Images}\label{Sec:Joint_train}

The above MIL-based WSIS method can be end-to-end optimized under box supervision. However, it is hard to perceive precise object boundary information. We propose to utilize a relatively small extra salient object segmentation dataset as auxiliary knowledge. These data that contain salient knowledge are jointly trained with box-supervised images to provide boundary localization guidance during the whole learning process.

Our joint training network for box-supervised class-agnostic segmentation (BoxCaseg) is shown in Fig.~\ref{fig:caseg}. The BoxCaseg network takes both salient images and box-supervised images as input. HRNet \cite{wang2020deep} is applied for feature extraction. The salient image feature and box-supervised image feature are fed into three segmentation heads supervised by the pixel-labeling loss and the MIL loss. The details are described in the following sections.


\vspace{-3mm}
\subsubsection{Salient Images}
\vspace{-2mm}
The salient images with mask annotations are derived from existing salient object segmentation/detection datasets, \eg, the DUTS-TR dataset \cite{wang2017learning}. In each salient image, we make sure there is only one object. The salient objects are from a large range of categories and occupy a large region in their images. The salient images are used for providing generalizable fine-grained boundary localization information. To avoid over-fitting, we make sure the salient images used in our method have no overlap with the target WSIS datasets, \eg, PASCAL and COCO. 

\vspace{-3mm}
\subsubsection{Data Augmentation}
\vspace{-2mm}
Different from BBTP which completes detection and segmentation simultaneously in Mask R-CNN, in which patch features are extracted using ROIAlign, we train a class-agnostic object segmentation model separately. Though ROIAlign is time-efficient to get patch features, it does not extract fine-grained features for precise object segmentation,  as discussed in PointRend \cite{kirillov2020pointrend}. Therefore, instead of using ROIAlign, we crop patches according to object positions as the input of our BoxCaseg network. The patch cropping (\ie, data augmentation) strategies are different for salient images and box-annotated images. For each salient image, it is firstly resized into $320\times 320$ and then randomly cropped into $288\times 288$ as the input. For box-annotated images, we find that the following more aggressive data augmentation strategy works better. 
\vspace{-1mm}
\begin{lstlisting}[language=Python]
def box_augmentation(x, y, w, h):
    x2 = x + random(-0.25, 0.25) * w
    y2 = y + random(-0.25, 0.25) * h
    w2 = random(0.5, 1.5) * w
    h2 = random(0.5, 1.5) * h
    return (x2, y2, w2, h2)
\end{lstlisting}    \vspace{-1mm}
This data augmentation strategy shifts the position and shape of the bounding box and introduces background information. Once a new box is obtained, an image patch is cropped and resized into $288\times 288$ for joint training.

\subsection{Multi-task Learning}
As shown in Fig.~\ref{fig:caseg}, there are three tasks for training our BoxCaseg network, \ie, MIL training for box-supervised images, MIL training for salient images, and pixel-labeling for salient images. There are three segmentation heads termed as salient segmentation head, transferred segmentation head, and weak segmentation head. The three heads have exactly the same network architecture as shown in the right side of Fig.~\ref{fig:caseg}.

\vspace{-3mm}
\subsubsection{MIL loss for weak segmentation}
\vspace{-2mm}
The convolutional feature for the salient image is denoted as $P_s$ and the convolutional feature for the weakly-supervised (\ie, box-supervised) image is denoted as $P_w$. $P_s$ and $P_w$ are fed into the weak segmentation head to obtain one-channel probability maps $S_s$ and $S_w$ via a pixel-wise sigmoid function, respectively. Besides, the upsample operation makes sure that $S_s$ and $S_w$ have the same size as the input images. As described in Sec.~\ref{sec:bbtp}, we can sample a set of positive bags $\mathbf{B}^+$ and a set of negative bags $\mathbf{B}^-$ from $S_s$ and $S_w$ respectively. The MIL loss is defined as follows.
\begin{equation}\label{eq:mil}
\begin{split}
    L_\text{mil}(S, \mathbf{B}^+, \mathbf{B}^-) = & -\sum_{B \in \mathbf{B}^+} \log \left( \max_{p \in B} S(p)  \right) \\
    & - \sum_{B \in \mathbf{B}^-} \log \left( 1 - \max_{p \in B}  S(p)  \right) \\
    & +  0.05\times\varphi(S, \mathbf{B}^+, \mathbf{B}^-) \text{,}
\end{split}
\end{equation}
where $p$ is a spatial position in a bag $B \in \{  \mathbf{B}^+, \mathbf{B}^- \}$ and a score map in $S \in \{S_s, S_w\}$, $S(p)$ denotes the score at position $p$, and the last item $\varphi(S, \mathbf{B}^+, \mathbf{B}^-)$ is a smooth term with a fixed weight of $0.05$ defined as follows. 
\begin{equation}\label{eq:smooth}
    \varphi(S, \mathbf{B}^+, \mathbf{B}^-) = \sum_{p\in \{ \mathbf{B}^+, \mathbf{B}^- \} } \sum_{p'\in{\Omega(p)}} {\|S(p)-S(p')\|}^2 \text{,}
\end{equation}
where $\Omega(p)$ means eight-connected region of the pixel $p$. In Eq.~\eqref{eq:mil}, the unary term uses MIL for pixel classification and the pairwise term aims to keep local smoothness. Positive bags having at least one pixel of the object, we encourage the maximal score of any positive bag to be as close to $1$ as possible, while suppressing maximal scores in negative bags.


\vspace{-3mm}
\subsubsection{Pixel-wise loss for salient segmentation and transferred segmentation}
\vspace{-2mm}
The pixel-wise loss function is only applied to salient images. The precise annotations of salient images with the pixel-wise loss are helpful for our BoxCaseg network to capture class-agnostic boundary information. As this loss cannot be applied to box-supervised images, we propose a transferred segmentation head to make the pixel-wise labeling parameters more suitable for box-supervised images. It is enabled by weight transfer, which is originally proposed in Mask$^{\star}$ RCNN \cite{hu2018learning}. As shown in Fig.~\ref{fig:caseg}, we transfer detached convolution parameters of the weak segmentation head to the transferred segmentation head via a weight transfer module. As for the weight transfer module, we simply use a two-layer multi-layer perceptron (MLP) with leaky ReLU as activation function. Note that the weight transfer is a single direction process, since it is detached from the weak segmentation head, \ie, the gradients do not back-propagate to the weak segmentation head. During training, the parameters of the weight transfer MLP is optimized. Given a salient image, its ground-truth is a binary mask denoted as $M$ and we denote its score maps output from the salient segmentation head and transferred segmentation head as $S_a$ and $S_t$ respectively. Then, the pixel-wise loss function is a standard binary cross-entropy loss defined as follows.
\begin{equation}\label{eq:pix}
\begin{split}
    L_\text{pix}(M, S_a, S_t) = &
    - M \log \left(  \alpha S_a + (1-\alpha) S_t  \right) \\ &
    - (1-M) \log \left( 1-  (\alpha S_a + (1-\alpha) S_t  \right)) \text{,}
\end{split}
\end{equation}
where $\alpha \in [0,1]$ is a parameter to control how much transferred segmentation map used for calculating the pixel-wise loss.

The final loss function is summarized as follows.
\begin{equation}\label{label}
     L = L_\text{mil}(S, \mathbf{B}^+, \mathbf{B}^-) + \lambda L_\text{pix}(M, S_a, S_t),
\end{equation}
where $\lambda$ is equal to $0$ if the input comes from box-supervised images, otherwise $1$. Last but not least, during the training of the proposed network, there are both box-supervised images and salient images in a mini-batch. Their ratios will be detailed in the experiments. This mixed input strategy helps stabilize the training process and solve the data imbalance problem, \eg, in the COCO dataset there are a huge number of box-supervised images while the number of salient images is limited.

\subsection{Training a Proxy Mask R-CNN}\label{sec:retraining}

To use jointly trained BoxCaseg network for instance segmentation, we need to obtain the positions of objects with a detector first in the testing phase. To simplify this process and avoid the time-consuming two-step testing procedure, we consider re-training a Mask R-CNN to perform box localization and mask prediction in one network. Our solution is straight-forward - we can use the BoxCaseg model to generate proxy masks on the training set. When generating proxy masks for training instances with the pre-trained BoxCaseg network, we just simply crop patches from the image by ground truth boxes as inputs and only use the salient segmentation head and transferred segmentation head to predict object mask. In our method, the Mask R-CNN model is trained with the bounding box annotations and the proxy masks, thus called Proxy Mask R-CNN. 

\begin{figure*}[htp]
\centering
\includegraphics[width=1\linewidth]{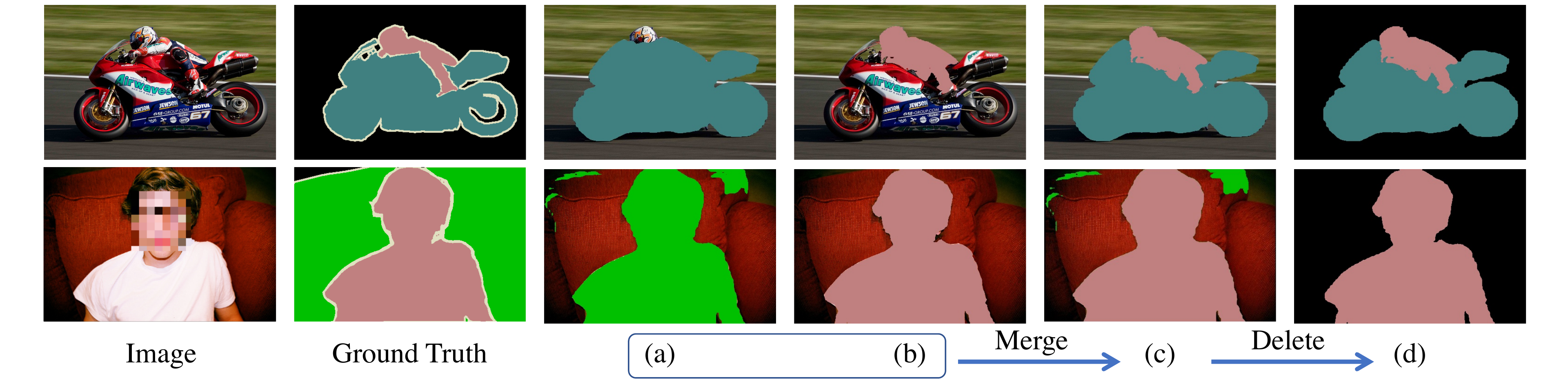}
\caption{From the prediction results of BoxCaseg (a\&b) to the proxy masks (d). Since there are two highly-overlapped objects in each image. BoxCaseg predicts two masks with errors. We first merge the two predicted masks using the \textit{smaller object better} strategy and then drop bad masks (\eg, the sofa mask in bottom row) according to the \textit{proxy box agreement} rule. In this case, the bounding box of the sofa mask in (c) has low agreement (\ie, low IoU) with its ground truth box, so the sofa mask is dropped.}
\label{fig:badcases}
\end{figure*}

Though the jointly trained BoxCaseg model obtains high performance, there are inevitably segmentation errors. The main difficulty is caused by the fact that multiple objects are highly-overlapped in one bounding box, \eg, motorbike with person and sofa with person in Fig.~\ref{fig:badcases}. Since these object pairs have large probabilities to appear together, weakly-supervised methods are hard to deal with them. To reduce the errors in the masks generated by BoxCaseg, we propose a \textit{merge and drop}, \ie, merge masks using the strategy of \textit{smaller object better} and drop masks via the \textit{proxy box agreement} rule, which are detailed as follows.

\textbf{Merging via smaller object better}:
Our prior knowledge tells us that each pixel in an image can only belong to one object. When merging segmentation results, for those pixels belong to multiple predicted masks, we assign them to the smallest objects. The reason is that the smaller object is usually within the box of the large one.



\textbf{Dropping via proxy box agreement}: After the merging procedure, we further drop low quality masks. It is observed that the low quality masks have shapes significantly different from their ground truth (GT), however, we only have GT boxes but do not have GT masks. Thus, we find the bounding box of the proxy mask, called proxy bounding box, and calculate the IoU between the proxy bounding box and GT bounding box as the proxy box agreement. It is the lower bound of mask IoU between proxy mask and GT mask and can reflect the segmentation quality. Note that dropping the mask does not mean removing it from the annotation file. Instead, those deleted masks are neglected in the back propagation process of mask head training in Mask R-CNN.

\vspace{-1mm}
\section{Experiment}

\paragraph{Datasets}

Extensive experiments are performed on the COCO 2017 dataset \cite{lin2014microsoft} and the augmented PASCAL VOC 2012 dataset \cite{hariharan2011semantic}. And we adopt the DUTS-TR dataset \cite{wang2017learning} to provide salient images.

\textbf{PASCAL VOC:} It is a commonly-used evaluating dataset for WSIS. Following the common practice \cite{khoreva2017simple}, augmented PASCAL VOC 2012 dataset is used, which is derived by combining the SBD dataset \cite{hariharan2011semantic} and the PASCAL VOC 2012 dataset \cite{everingham2015pascal}. The final dataset contains $10,582$ training images with $25,815$ training instances and $1,449$ validation images.

\textbf{COCO:} It is a large-scale instance segmentation dataset which contains $118,287$ training images with $860,001$ instances and $5,000$ images for validation. It is also applied for evaluating WSIS in our experiments.

\textbf{DUTS-Single:} The images in the DUTS-TR dataset \cite{wang2017learning} are collected from ImageNet DET \cite{deng2009imagenet} training/val/test sets and the SUN dataset \cite{5539970}. DUTS-TR is widely-used for salient object segmentation/detection task. \textit{We have carefully checked the image sources of DUTS-TR, they have no overlap with Pascal VOC and COCO dataset.} Usually, a bounding box contains only one object, so we filter out images which contain multiple objects to ensure that the foreground mask only belongs to a single object. The derived salient object segmentation dataset is named  DUTS-Single, which contains $7991$ images. The derived DUTS-Single dataset will be made publicly available.


\vspace{-3mm}
\paragraph{Evaluation metric}

For evaluating the performance of class-agnostic object segmentation, we use IoU@\emph{k}, which represents the percentage of predicted instance masks whose intersection over union (IoU) with its ground-truth mask is larger than \emph{k}\%, where \emph{k}$\in$\{50, 75\} in our experiments. Besides, we design a mIoU$^{*}$ metric to evaluate our model's class-related performance, which calculates IoUs of every instance prediction, per-class mean IoU and averages all the per-class mean IoUs.



After retraining the Proxy Mask R-CNN, we adopt the standard evaluation metric, the mean average precision (mAP). We follow the same evaluation scheme as \cite{hsu2019weakly} and report mAP with four IoU thresholds (including 0.25, 0.50, 0.70 and 0.75) for PASCAL VOC dataset. Besides, we also report the evaluation metric including AP, AP50, AP75 (averaged over IoU thresholds), and APS, APM, APL (AP at different scales) for COCO dataset.



\vspace{-3mm}
\paragraph{Implementation details}
\vspace{-1mm}

For joint training the BoxCaseg network, we apply the HRNetV2-W48 \cite{wang2020deep} as the backbone and its default hyper-parameters are applied, \ie, initial learning rate of 4e-3, weight decay of 1e-4, momentum of 0.9. The batch-size is set to be 16 per GPU. For generating proxy masks for training Mask R-CNN, each training object is cropped from its image according to its ground-truth bounding box and resized into 320$\times$320 as the input of pre-trained BoxCaseg network. Proxy Mask R-CNN is trained using ResNet-101-FPN \cite{he2016deep} as the backbone following the default settings in the maskrcnn benchmark \cite{massa2018mrcnn}. Experiments are carried out on two Nvidia TITAN V GPUs (12GB RAM). Source code and pre-trained models will be made publicly available.


\vspace{-1mm}
\subsection{Comparison with the State-of-the-art Methods}
\vspace{-1mm}

Tab.~\ref{tab:pascal} presents the instance segmentation results on the PASCAL VOC 2012 validation set, including the fully-supervised Mask R-CNN baseline, weakly-supervised methods using image-level annotation and box-level annotation. In the metrics, $AP_{50}$ is the most widely used one. Our $AP_{50}$ is 67.6, which is on par with the fully-supervised Mask R-CNN. With the help of joint training with salient images, our method significantly outperforms other weakly-supervised instance segmentation methods. Comparing with the MIL-based BBTP method \cite{hsu2019weakly}, joint training with salient images brings 8.7 $AP_{50}$ performance gain and 20.8 $AP_{75}$ performance gain. The results confirm that joint training with salient images is very helpful for precisely segmenting objects.

\begin{table}[htp]
    \caption{Instance segmentation results on the PASCAL VOC 2012 val set. In the supervision column, M denotes mask-level supervision, I denotes image-level supervision, B denotes box-level supervision and S denotes salient object supervision. $^\ddag$ indicates the results are obtained by running their released codes using our settings.} 
    \centering
    \resizebox{1\linewidth}{!}{
    \begin{tabular}{lccccc}
        \toprule
        Method & Sup. & $AP_{25}$ & $AP_{50}$ & $AP_{70}$ & $AP_{75}$ \\
        \midrule
        \multicolumn{6}{c}{ResNet-50 Backbone}\\
        \hline
        PRM \cite{zhou2018weakly} & I & 44.3 & 26.8 & - & 9.0 \\
        IAM \cite{zhu2019learning}& I & 45.9 & 28.8 & - & 11.9 \\
        Label-PEnet \cite{ge2019label}& I & 49.1 & 30.2 & - & 12.9 \\
        CountSeg \cite{cholakkal2019object}& I  & 48.5 & 30.2 & - & 14.4 \\
        WISE \cite{laradji2019masks}& I & 49.2 & 41.7 & - & 23.7 \\
        IRN \cite{ahn2019weakly}& I & - & 46.7 & 23.5 & - \\
        LIID \cite{liu2020leveraging} & I & - & 48.4 & - & 24.9 \\
        BBTP$^\ddag$ \cite{hsu2019weakly} & B+S & {75.9} & {61.0} & {33.4} & {25.1} \\ 
        ShapeProp$^\ddag$ \cite{zhou2020learning} & B+S & {75.1} & {61.6} & {37.8} & {30.3} \\ 
        BoxCaseg (ours) & B+S & \textbf{76.2} & \textbf{65.8} & \textbf{46.3} & \textbf{38.6} \\ 
        \midrule
         \multicolumn{6}{c}{ResNet-101 Backbone}\\
         \hline
        SDI \cite{khoreva2017simple}& B  & - & 46.4 & - & 18.5 \\
        AnnoCoIn \cite{arun2020weakly}& B & 73.1 & 57.7 & 33.5 & 31.2 \\    
        BBAM \cite{bbam} & B & 76.8 & 63.7 & 39.5 & 31.8 \\     
        BoxInst \cite{tian2020boxinst} & B & - & 61.4 & - & 37.0 \\   
        BBTP \cite{hsu2019weakly}& B  & 75.0 & 58.9 & 30.4 & 21.6 \\  
        BBTP$^\ddag$ \cite{hsu2019weakly}& B+S& 77.0 & 61.4 & 34.4 & 26.6 \\  
        ShapeProp$^\ddag$ \cite{zhou2020learning} & B+S& 76.0 & 62.4 & 39.9 & 31.3 \\  
        BoxCaseg (ours) & B+S & \textbf{77.7} & \textbf{67.6} & \textbf{49.4} & \textbf{42.4} \\ 
        \midrule
        Mask R-CNN \cite{he2017mask} & M & \textcolor{gray}{76.7} & \textcolor{gray}{67.9} & \textcolor{gray}{52.5} & \textcolor{gray}{44.9}\\
        \bottomrule
    \end{tabular}
    }
    \label{tab:pascal}
\end{table}

Tab.~\ref{tab:coco} presents the results on the COCO test-dev set. COCO is a large-scale instance segmentation dataset that contains $118,287$ training images that contain about $860,000$ training instances. Using only $7,991$ (9.3\%) salient objects for joint training, we improve the state-of-the-art baseline (BBTP) by 10.1 $AP$ and the results are getting close to the fully-supervised Mask R-CNN.

\begin{table*}[htp]
    \caption{Instance segmentation results on the COCO test-dev set.} 
    \centering
    \resizebox{0.86\linewidth}{!}{
    \begin{tabular}{lcccccccc}
        \toprule
        Method & Supervision & Backbone & $AP$ & $AP_{50}$ & $AP_{75}$ & $AP_{S}$ & $AP_{M}$ & $AP_{L}$ \\
        \midrule
        Mask R-CNN \cite{he2017mask} (Oracle) & M & ResNet-101-FPN  & \textcolor{gray}{35.7} & \textcolor{gray}{58.0} & \textcolor{gray}{37.8} & \textcolor{gray}{15.5 }& \textcolor{gray}{38.1} & \textcolor{gray}{52.4} \\  
        BBTP \cite{hsu2019weakly} & B & ResNet-101-FPN   & 20.8 & 45.1 & 17.0 & 10.4 & 21.7 & 30.3 \\
        BBAM \cite{bbam} & B & ResNet-101-FPN   & 25.7 & 50.0 & 23.3 & - & - & - \\
        BoxCaseg (ours) & B+S & ResNet-101-FPN   & \textbf{30.9} & \textbf{54.3} & \textbf{30.8} & \textbf{12.1} & \textbf{32.8} & \textbf{46.3} \\   
        \bottomrule
    \end{tabular}
    }
    \label{tab:coco}
\end{table*}

\begin{table}[htp]
    \caption{Results on PASCAL val using different numbers of salient images for joint training.} 
    \centering
    \resizebox{1\linewidth}{!}{
    \begin{tabular}{c|ccccc}
        \toprule
        Number        & mIoU* & IoU@50 & IoU@75 & AP@50 & AP@75 \\
        \midrule
        100             & 61.8  & 70.4  & {32.2} & 55.3 & 27.0 \\
        500             & 65.3  & 73.3  & {39.6}  & 62.6 & 36.2 \\
        1000            & 66.4  & 77.4  & 47.3 & 65.1 & 37.2 \\ 
        2500            & 69.2  & 79.9  & 51.1 & 67.1 & 40.9 \\
        {4701}            & 70.1 & 80.1  & 50.2 & 66.7 & 38.9 \\
        All             & 71.8  & 83.8  & 56.8 & 67.6 & 42.4 \\
        \bottomrule
    \end{tabular}}
    \label{tab:lessalient}
\end{table}

\vspace{-1mm}
\subsection{Ablation Studies}
\vspace{-1mm}

To validate our proposed method and configurations' usefulness, extensive ablation experiments are conducted on both the joint training BoxCaseg stage and the re-training stage. 

\vspace{-3mm}
\subsubsection{Less salient images for joint training}
\vspace{-1mm}

In order to investigate the influence of the number of salient images, we perform ablation experiments on the number of salient images in $\{100, 500, 1000, 2500, 4701, 7991/\text{All}\}$ for joint training. The $100, 500, 1000$, \& $2500$ salient images are randomly selected. \textit{The $4701$ salient images are manually picked to make sure that they do not contain any object belongs to any of the $20$ PASCAL categories.}

As Tab.~\ref{tab:lessalient} shows, when there are too few salient images ($100$), our salient segmentation head and transferred segmentation head will perform awful and serve unstable performance, even worse than the performance of weak segmentation head. When the number of salient images comes to over $500$, its training process becomes stable and the model will perform better with images number increase both in mIoU$^{*}$ and AP evaluation metric. Training with the $4701$ salient images, the results are very close to that uses all salient images, which confirms that BoxCaseg has a very strong category-wise transfer ability.



\vspace{-3mm}
\subsubsection{The effectiveness of the three segmentation heads}
\vspace{-1mm}

Tab.~\ref{tab:segheads} indicates the influence of three segmentation heads for the whole model respectively. First, only using a salient segmentation head, which is trained by salient images, will lead to an awful segmentation result on box-supervised images. This is due to the salient images are different from the box-supervised PASCAL images. So it is necessary to add the weak segmentation head to help the backbone learn how to extract features of box-supervised images. Second, transferred segmentation head, which delivers box-supervised information from the weak segmentation head to pixel-level prediction, helps the salient segmentation head improve segmentation quality on box-supervised images. Finally, by using all three heads, we finally get a $71.8$ mIoU$^{*}$ instance segmentation result on Pascal VOC 2012 validation set. In fact, our BoxCaseg model has a strong salient object segmentation performance that is on-par with the SOTA U$^2$Net\cite{qin2020u2}.



\begin{table}[htbp]
    \caption{Results on PASCAL val when using different segmentation heads. Weak, Salient, and Trans denote the weak, salient, and transferred segmentation heads respectively. If only Trans or Salient is checked, $\alpha$ in Eq.~\eqref{eq:pix} is set to $0$ or $1$.}
    \centering
    \resizebox{0.93\linewidth}{!}{
    \begin{tabular}{ccc|ccc}
        \toprule
        Weak & Salient & Trans. & mIoU$^{*}$  & IoU@50 & IoU@75 \\
        \midrule
        \checkmark & &                           & 66.1 & 76.8 & 40.0 \\
         & \checkmark &                          & 43.1 & 43.0 & 25.9 \\
        \checkmark &            & \checkmark     & 69.7 & 80.0 & 50.3 \\
        \checkmark & \checkmark &                & 70.5 & 80.8 & 53.7 \\
        \checkmark & \checkmark & \checkmark     & \textbf{71.8} & \textbf{83.4} & \textbf{56.8} \\
        \bottomrule
    \end{tabular}
    }
    \label{tab:segheads}
\end{table}

\vspace{-3mm}
\subsubsection{Sampling strategies for imbalanced training images}
\vspace{-1mm}

When there is an extreme unbalance between COCO ($860,001$ instances) and DUTS-Single ($7991$ instances), we discover that the model performs worse as the training progressing. Thus, we propose a simple yet efficient sample strategy that we use all salient images and randomly sample part of box-supervised images to form batches in an epoch. And all batches own a common fixed ratio of box-supervised images to salient images. Tab.~\ref{tab:COCOSampleVal} demonstrates our sample strategy is efficient, where $r_{ws}$ is the ratio of box-supervised images to salient images in a batch and RS represents random sampling. Obviously, this strategy greatly solves the problem of data unbalance and makes training process more stable as well as enables model with better generalization ability. When $r_{ws}$ is equal to $9:7$, we can get $70.9$ mIoU$^{*}$ in COCO validation set.


\begin{table}[htbp]
    \caption{Results on COCO val using different image sampling methods. RS means randomly sampling from the union set of box-supervised images and salient images. $r_{ws}$ is the ratio of box-supervised images to salient images in a batch.}
    \centering
    \resizebox{0.68\linewidth}{!}
    {
    \begin{tabular}{c|ccc}
        \toprule
        $r_{ws}$ & mIoU$^{*}$  & IoU@50 & IoU@75 \\
        \midrule
        RS            & 58.1 & 62.2 & 27.9 \\
        8:8           & 70.7 & 84.5 & 48.0 \\
        9:7           & \textbf{70.9} & \textbf{84.9} & \textbf{48.6} \\
        10:6          & 70.0 & 82.7 & 45.4 \\
        11:5          & 70.7 & 83.6 & 47.8 \\
        12:4          & 70.4 & 82.7 & 47.2 \\
        13:3          & 68.1 & 77.6 & 44.1 \\
        14:2          & 69.2 & 81.1 & 46.4 \\
        \bottomrule
    \end{tabular}
    }
    \label{tab:COCOSampleVal}
\end{table}

\vspace{-3mm}
\subsubsection{The weight of transferred segmentation head}
\vspace{-1mm}

We also conduct experiments on $\alpha$, which controls the weight of the transferred segmentation head whn adding with the salient segmentation head for pixel-level prediction. As Tab.~\ref{tab:ablation_alpha} shows, combining the prediction of transferred segmentation head and salient segmentation head, we have access to deriving better results than only using one of two heads by at least $1.3$\% improvement. What's more, it's relatively stable for segmentation result, when $\alpha \in [0.3, 0.7]$ and proves our weight transfer method's usefulness. And we can derive the best result when $\alpha$ is $0.7$.

\begin{table}[htbp]
    
\caption{Results on PASCAL val set when the weight of salient segmentation head ($\alpha$) varies.}
    \resizebox{1\linewidth}{!}{
    \centering
    {
    \begin{tabular}{c|ccccccc}
        \toprule
        $\alpha$ & 0.0 & 0.3 & 0.4 & 0.5 & 0.6 & 0.7 & 1.0 \\
        \midrule
        mIoU$^{*}$   & 69.7 & 71.7 & 71.0 & 71.4 & 71.3 & \textbf{71.8} & 70.5\\
        IoU@50     & 80.0 & 83.0 & 82.0 & 82.6 & 82.7 & \textbf{83.4} & 80.8 \\
        IoU@75     & 50.3 & 55.7 & 54.9 & 55.6 & 55.2 & \textbf{56.8} & 53.7 \\
        \bottomrule
    \end{tabular}
    }
    }
    \label{tab:ablation_alpha}
\end{table}


\begin{table}[htbp]
    \caption{Results on PASCAL val and COCO val when using the merging and dropping strategies or not.} 
    \centering
    \resizebox{0.85\linewidth}{!}{
    \begin{tabular}{ccc|ccc}
        \toprule
        Data & Merge & Drop & $AP$ & $AP_{50}$ & $AP_{75}$ \\
        \midrule
        PASCAL & & & 39.6 & 67.5 & 41.3  \\
        PASCAL & \checkmark &  & \textbf{40.2} & \textbf{67.6} & 42.1 \\
        PASCAL & \checkmark & \checkmark & \textbf{40.2} & \textbf{67.6} & \textbf{42.4}  \\    
        \midrule
        COCO & & & 28.6 & 51.3 & 28.2  \\
        COCO & \checkmark & & 29.5 & 51.9 & 29.3  \\
        COCO & \checkmark & \checkmark & \textbf{30.9} & \textbf{53.7} & \textbf{31.3} \\
        \bottomrule
    \end{tabular}
    }
    \label{tab:retrainstrategies}
\end{table}

\vspace{-3mm}
\subsubsection{The strategies for merging and deleting proxy masks}
\vspace{-1mm}

In Tab.~\ref{tab:retrainstrategies}, we firstly give a baseline of retraining results without any strategies. We can observe that the merge via smaller object better strategy improves the performance by $0.8$ $AP_{75}$ and $1.1$ $AP_{75}$ on the two datasets respectively. At the same time, the drop via proxy box agreement strategy obtains obvious improvement on the COCO dataset while little improvement on the PASCAL dataset. This is due to the fact that PASCAL is simpler than COCO dataset and bad cases are much fewer. In addition, we tried different values for the dropping threshold to find the best performing threshold, shown in Tab.~\ref{tab:dropratio}. The drop rate refers to the percentage of the ignored samples in the total samples. The results show that the drop strategy can pick out samples with better proxy masks and bring robust performance improvements. Using a very high IoU threshold of $0.95$ obtains the best results among them, which reveals that have a few very high quality mask for training Mask R-CNN is better than having lots of medium quality masks.

\begin{table}[htbp]
\caption{COCO val results and the ratios of dropped masks when the dropping threshold varies.}
  \centering
  \resizebox{0.86\linewidth}{!}{
  \begin{tabular}{l|c c c c c }
  \toprule
    Drop thresh & 0.0 & 0.75 & 0.85 & 0.90 & 0.95 \\
    \midrule
    Drop rate & 0.00 & 0.27 & 0.49 & 0.64 & 0.80 \\
    COCO AP & 29.5 & 30.2 & 30.4 & 30.7 & \textbf{30.9} \\
  \bottomrule
  \end{tabular}
  }
\label{tab:dropratio}
\end{table}

\vspace{-3mm}
\section{Conclusion}
\vspace{-1mm}

We explore a new direction for WSIS, \ie, box-supervised class-agnostic learning with salient images for object segmentation. The proposed BoxCaseg network uses the salient object information as an auxiliary memory during the whole training process. This design is to mimic the way humans perform class-agnostic object segmentation. Once the proxy masks are obtained via the pretrained BoxCaseg model, we propose useful mask merging and dropping strategies for re-training Mask R-CNN. On PASCAL and COCO, state-of-the-art weakly-supervised instance segmentation results have been achieved. In the next step, we would like to develop new learning methods to push the performance of the BoxCaseg-based WSIS method to match the performance of fully-supervised Mask R-CNN on COCO using the $7991$ salient images.

\vspace{-3mm}
{\small \paragraph{Acknowledgement:} This work was in part supported by NSFC (No. 61876212 and No. 61733007) and the Zhejiang Laboratory under Grant 2019NB0AB02.}

{\small
\bibliographystyle{ieee_fullname}
\bibliography{main}
}

\end{document}